\documentclass[letterpaper, 10 pt, conference]{ieeeconf}  

\IEEEoverridecommandlockouts                              

\usepackage{times}
\usepackage{graphicx,stfloats}
\DeclareGraphicsExtensions{.pdf,.png,.jpg}
\usepackage{amsmath}
\DeclareMathOperator*{\argmin}{argmin} 
\usepackage{amssymb}  
\usepackage{makecell}
\usepackage{booktabs}
\usepackage{multirow}
\usepackage{multicol}
\usepackage{here}
\usepackage{kotex}
\usepackage{soul, color}
\usepackage{subcaption} 
\usepackage{anyfontsize} 
\usepackage{hyperref}
\usepackage{booktabs}  
\usepackage{cuted}
\usepackage[font=small,labelfont=bf,tableposition=top]{caption}
\usepackage{etoolbox}

\title{\LARGE \bf
 Just Flip: Flipped Observation Generation and Optimization for Neural Radiance Fields to Cover Unobserved View
}

\author{Minjae Lee, Kyeongsu Kang and Hyeonwoo Yu
 \thanks{Minjae Lee, Kyeongsu Kang and Hyeonwoo Yu are with the Department of Electrical Engineering \& Graduate School of AI, Ulsan National Institute of Science and Technology (UNIST), Ulsan, South Korea. {\tt\small \{lmjbsj, thithin, hyeonwoo.yu\}@unist.ac.kr}}
 \thanks{
 The code will be available at: \href{https://github.com/minjae-lulu/Just-Flip}{https://github.com/minjae-lulu/Just-Flip}} %
}

\begin{document}



\maketitle
\vspace*{-10cm}
  \begin{strip}
    \centering
    \includegraphics[width=17.8cm, height=5.0cm]{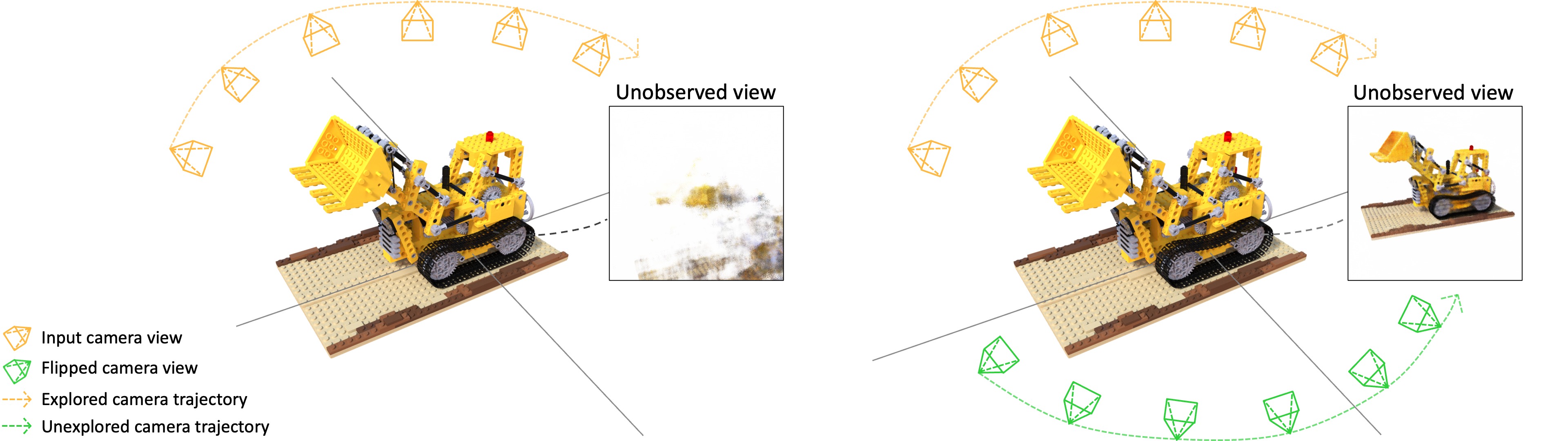}
    \captionof{figure}{Overview of our method. 
       (Left) The baseline approach where the robot only observes one side of an object while driving. This case does not yield good rendering results in unobserved views that the robot has not explored. (Right) Our method generates the flipped observations from the actual observations. The robot exploits both input images and flipped images and estimated camera poses to learn 3D space using NeRF for unexplored regions as well. Our method obtains qualified rendering results in unobserved views, even without providing images from unobserved views as a training set.}
       \label{overview_image}
  \end{strip}


\begin{abstract}


With the advent of Neural Radiance Field (NeRF), representing 3D scenes through multiple observations has shown significant improvements.
Since this cutting-edge technique is can obtain high-resolution renderings by interpolating dense 3D environments, various approaches have been proposed to apply NeRF for the spatial understanding of robot perception.
However, previous works are challenging to represent unobserved scenes or views on the unexplored robot trajectory, as these works do not take into account 3D reconstruction without observation information.
To overcome this problem, we propose a method to generate flipped observation in order to cover unexisting observation for unexplored robot trajectory.
To achieve this, we propose a data augmentation method for 3D reconstruction using NeRF by flipping observed images, and estimating flipped camera 6DOF poses. 
Furthermore, to ensure the NeRF model operates robustly in general scenarios, we also propose a training method that adjusts the flipped pose and considers the uncertainty in flipped images accordingly. Our technique does not utilize an additional network, making it simple but fast, thereby making it suitable for robotic applications where real-time performance is important. We demonstrate that our method significantly improves three representative perceptual quality measures on the NeRF dataset.

\end{abstract}

\section{Introduction}







Simultaneous Localization and Mapping (SLAM) is one of the key methods both in the fields of robotics and computer vision. SLAM is the process of generating a 3D map of an unknown environment while simultaneously estimating the position and movement of a robot or camera. This technique is being actively researched due to its applications in various real-world scenarios such as autonomous vehicles, unmanned robots, etc \cite{filipenko2018comparison, yu2018ds, yu2019variational, mittal2022vision}.
Recently, in computer graphics, a novel technology called Implicit Neural Representation (INR) has emerged that utilizes neural networks to parameterize continuous and differentiable signals. As one of the INR, Neural Radiance Field (NeRF) \cite{mildenhall2021nerf} is a deep learning-based approach for 3D scene representation. In this approach, network learns 3D space by projecting a set of 2D views of a scene into a continuous 3D space, such that the color and density information from each view, thus new views of the scene is interpolated for 2D rendering.
This technique can be applied to 3D mapping for SLAM as it can be exploited as a 3D scene representation.
It also has several advantages over existing SLAM systems in that it can represent space continuously and is more memory-efficient as it utilizes neural networks to represent space. Due to these features, many recent works \cite{sucar2021imap, zhu2022nice, rosinol2022nerf, yang2022vox, kong2023vmap} have applied NeRF to SLAM and 3D mapping.

However, despite many advantages of NeRF, similar to the previous 3D perception methods it also has limitations in synthesizing unobserved views; from here, we define an unobserved view as a view that the robot has not explored or cannot explore. We also define the unobserved view problem as the task of estimating the synthesized view of an unexplored area.
For example, in Fig.~\ref{overview_image}, the robot explores the orange camera trajectory while observing the lego truck, and the green cameras and trajectory represent unobserved views and the unexplored trajectory respectively. 
The view from the green cameras is unobserved views because they have not been explored yet.
These unobserved views cannot be obtained from the observation trajectory due to the 2.5D observation of the robot and the self-occlusions of the objects.
Since it is challenging to obtain observations of the unexplored region, various approaches based on NeRF show poor rendering performance on the unexplored scene.
Previous research such as \cite{xiang2015data} tried to overcome the occlusion problem by capturing visibility patterns of 3D voxel exemplars, but they are learning-based methods which require a massive training dataset.
The robot needs to be adaptive in environments where it encounters various objects, so NeRF achieves some advantages compared to the method based on the object dataset \cite{xiang2015data} while it does not require pretraining in such challenging environments.
However, as mentioned above, the robot is hard to achieve unobserved views while navigating and thus obtains limited input image data.
Therefore, training NeRF with a few images becomes important and various data augmentation methods such as \cite{chen2022geoaug, bortolon2022data} have been studied.
Although these methods use geometric approaches to acquire views between or near explored views, they still have limitations in that they do not consider any information about unexplored areas, resulting in the unobserved view problem. Moreover, since they require model training for data augmentation, they may not be suitable for robotics applications that require real-time processing.

To address the unobserved view problem in NeRF, we propose a new unobserved view generation and optimization method. We assume that artificially created objects in indoor and outdoor environments mostly exhibit symmetrical 3D shapes \cite{tulsiani2015shape, mitra2013symmetry, wu2020unsupervised}. Typical representative datasets also exhibit symmetric characteristics of 3D objects.
Under this assumption, by flipping object observation we can generate additional significant information of unobserved views. Since training NeRF requires both the images and camera poses, estimating the camera position for the flipped observation image is also imperative.
Fortunately, some works \cite{yen2021inerf, wang2021nerf, meng2021gnerf, lin2021barf, xu2022sinnerf, bian2023nope} that try to train NeRF without camera poses have been introduced.
However, these methods cannot guarantee finding the global minimum of the camera poses where the camera pose is completely randomly initialized without additional model. To relax this problem of our method, we propose a method to estimate the 6DOF camera pose of the flipped image using the camera pose of the input image.
But, this flip and estimation method is not effective for complex objects. To adapt to objects of various shapes, we refine the flipped camera pose and employ a Bayesian approach that incorporates the uncertainty of the flipped image. Therefore we show that it is possible to exploit the symmetric parts of the given 3D object and ignore the unnecessary part according to the estimated uncertainty, thereby we have the robust unobserved view prediction approach.

In summary, we propose a flipped observation generation and training method by considering robot trajectory to improve understanding of unobserved views. 
Our contribution is as follows:
\begin{itemize}
    \item 
    We introduce the flip method for predicting unobserved views and propose an approach for the initial 6DOF pose estimation of the flipped images.

    \item 
    Recognizing that the initialized flipped camera pose can be noisy depending on the object's shape, we propose a learning approach that performs bundle adjustment on the flipped pose. This method simultaneously optimizes the camera pose and network parameters.

    \item 
    Flipped images also inherently have uncertainties depending on the object. To address this, we use Bayesian approach that integrates these uncertainties into the model's loss during the learning process. \\

\end{itemize}

\section{Related work} 

In the traditional SLAM techniques, 3D representation was achieved using methods such as mesh \cite{riegler2020free, riegler2021stable}, point clouds \cite{qi2017pointnet, achlioptas2018learning}, and depth maps \cite{liu2015learning, huang2018deepmvs}. However, these approaches have the drawback of having memory limitations since they require storing discrete information, resulting in the map being represented in a sparse manner. To address these limitations, recent research \cite{sucar2021imap, zhu2022nice, rosinol2022nerf, yang2022vox, kong2023vmap} has applied NeRF to SLAM for 3D mapping. NeRF defined as a continuous function by a neural network that takes in a spatial location and viewing direction as inputs and outputs the RGB values and volume density. By continuous neural network function, their approach has the advantages of high resolution and low memory. NeRF has shown remarkable performance in synthesizing photorealistic novel views of real-world scenes or objects. Afterward, NeRF have been extended to achieve fast training \cite{chen2021mvsnerf, xu2022point, muller2022instant}, few input data \cite{yu2021pixelnerf, deng2022depth}, large scale scene \cite{xiangli2021citynerf, tancik2022block, turki2022mega}, dynamic scene capture \cite{park2021nerfies, yang2022banmo}, scene editing \cite{wang2022clip, yuan2022nerf}.

In neural radiance fields, we attempt to effectively predict unobserved views in a few-shot setting for robots by estimating the camera pose and uncertainty associated with flipped images. Since the flipped image does not have a camera pose, NeRF needs to be trained without camera pose or simultaneously estimate the camera pose and neural radiance fields. This challenge has been tackled in some works. \cite{yen2021inerf} implemented pose estimation by inverting a neural radiance field, while \cite{wang2021nerf} adopted a two-stage strategy optimizing both camera parameters and the radiance field. With the use of the SIREN layer, \cite{xu2022sinnerf} introduced sampling methods to counter the joint optimization in NeRF. However, they can only optimize camera pose for relatively short camera trajectories. \cite{meng2021gnerf} can reconstruct neural radiance fields and estimate camera poses using generative model. \cite{lin2021barf} applies bundle adjustment, and \cite{bian2023nope} utilizes monocular depth estimation, have also been explored. Amidst this backdrop and given our approach to flipped images, the task of addressing inherent uncertainty becomes crucial. There are several prior studies that have incorporated uncertainty into NeRF. Specifically, \cite{martin2021nerf} introduced the concept of uncertainty to render complex scenes from unstructured images. In this approach, they derive uncertainty from transient networks and consider it in the pixel levels of the images. 
\cite{shen2021stochastic} produce uncertainty estimates by modeling a distribution over all the possible radiance fields modeling the scene. They represent the distribution of all potential radiance fields for scene modeling using variational inference. \cite{shen2022conditional} by the same authors address the limitations of \cite{shen2021stochastic} by integrating a conditional normalizing flow with latent variable modeling. Based on these methods, we introduce a flip data generation and training method within NeRF to tackle the unobserved view problem. \\

\section{Preliminary}





NeRF is a deep learning-based approach for 3D scene reconstruction from 2D images. It aims to model a 3D scene as a continuous function that maps a 3D point in space to its color and intensity in the image plane. This function is represented as a neural network, which is trained on a set of input images $\mathcal{I} = \{I_1, I_2, ... , I_N\}$ of a scene, with their associated camera parameters $\Pi = \{\pi_1, \pi_2, ... , \pi_N\}$. NeRF model is to learn a mapping of image radiance $\mathbf{c} = (r,g,b)$ and spatial intensity $\sigma$ for 3D coordinates $\mathcal{X}=\{\mathbf{x}\}$ at viewing direction $\mathbf{d} = (\theta, \phi)$. This mapping is represented by a neural network and the mapping function can be represented mathematically as $F_\Theta : (\mathbf{x},\mathbf{d}) \rightarrow (\mathbf{c}, \sigma)$, where $\Theta$ is network parameters.

In order to obtain the rendered image $\hat{I_i}$, the color at each pixel $\mathbf{p} = (u,v)$ is estimated by a rendering function $\mathcal{R}$. The expected image color $\hat{I_i}(\mathbf{p})$ of camera ray $\mathbf{r}(t) = \mathbf{o} + t\mathbf{d}$ that starts from camera origin $\mathbf{o}$ with near and far bound $t_n$ and $t_f$ can be written as:
\begin{align}
    \nonumber
    \hat{I_i}(\mathbf{p}) &= \mathcal{R}(\mathbf{p}, \pi_i|\Theta) \\
    &=  \int_{t_n}^{t_f} T(t)\sigma(\mathbf{r}(t))\mathbf{c}(\mathbf{r}(t),\mathbf{d})dt\text{,}
    \label{eqn:rendering function}
\end{align}
where
\begin{align}
     \nonumber \\
    T(t) = exp(-\int_{t_n}^{t}\sigma(r(s))ds) \text{.}  
\end{align}
$T(t)$ denotes the accumulated transmittance along the ray from $t_n$ to t, \textit{i.e.}, the probability that the ray travels from $t_n$ to $t$ without hitting any other particle. Therefore, the network can be trained by minimizing the difference between the predicted (or rendered) image $\hat{I}$ and the ground truth input image $I$. This difference can be defined as photometric loss $L_p$:
\begin{align} 
    \label{eqn:nerf_loss}
    L_p(\Theta, \Pi) = \sum_i^N \|I_i - \hat{I_i}\|_2^2 \text{.} 
\end{align}
\\
\section{Approach}




\subsection{NeRF with Camera Pose}
\subsubsection{Initial Camera Pose}
Symmetry is prevalent in real-world objects due to its functional and aesthetic benefits, and can be manufactured easily in some cases \cite{tulsiani2015shape, mitra2013symmetry, wu2020unsupervised}. Based on this premise, our idea starts flipping the input images $\{I\}$ in order to predict unobserved views for the unexplored area. 
Then Eqn.~\eqref{eqn:nerf_loss} can be written as follows:
\begin{align} 
    \label{eqn:redefined_nerf_loss}
    L_p(\Theta, \Pi^{all}) = \sum_i^N \|I_i - \hat{I_i}\|_2^2 + \sum_i^N \|I_i^{\prime} - \hat{I_i^{\prime}}\|_2^2 \text{,}
\end{align}
where $\{I_i^{\prime}, ... , I_n^{\prime}\}=\mathcal{I^{\prime}}$ are flipped images and $\Pi^{all}$ is a union set of the existing camera poses $\Pi=\{\pi_i, ... , \pi_n\}$ of $\mathcal{I}$ and camera poses $\Pi^{\prime}=\{\pi_i^{\prime}, ... , \pi_n^{\prime}\}$ for $\mathcal{I^{\prime}}$.
In this case, a challenge arises since the flipped images $\{I^\prime\}$ for the unobserved views do not possess defined camera poses.
To relax this issue, approaches such as \cite{schonberger2016structure, lin2021barf} based on Bundle adjustment (BA) can be adopted. However, those BA-based approaches require a larger number of images than those acquired from scenarios such as exploration or navigation, or require a nice initial guess for the camera poses. Thus, it is not suitable to directly apply those approaches.

Thus, we propose a method to estimate the initial camera pose for the flipped images $\{I^\prime\}$ by leveraging the existing camera pose and geometric constraints.
This method commences by using the least squares approach to identify the optimal sphere that passes through the given input camera pose.
The general equation of a sphere with its center at $x_0, y_0, z_0$ and radius $r$ is represented by $x^2+y^2+z^2=2 x x_0+2 y y_0+2 z z_0+r^2-x_0^2-y_0^2-z_0^2$.
Using input camera poses, we can express the equation of a sphere in matrix form as $\vec{f} = A\vec{c}$, where $\vec{c}$ contains information regarding the radius and the center coordinates of the sphere.
Through the matrix equation, we can find the value of $\vec{c}^\prime$ that minimizes the norm of the residual as the following:
\begin{align} 
    \nonumber 
    \vec{c}^\prime =& \argmin_{\vec{c}} (A\vec{c} - \vec{f})^T (A\vec{c} - \vec{f})  \text{.} 
\end{align}
Once the optimized sphere is obtained, the input camera pose coordinates $x,y,z$ undergo a symmetrical transformation through the symmetric plane that intersects the center of the sphere. Subsequently, leveraging the transformed camera coordinates $\mathbf{c} = (x^{\prime}, y^{\prime}, z^{\prime})$, the sphere's center point $\mathbf{at} = (x_0, y_0, z_0)$, and the up vector  $\mathbf{up} = (0, 0, 1)$, we can compute the rotation matrix $R$.
Naturally, transformed camera pose corresponds to the translation vector $T$ and by integrating it with the rotation matrix $R$, we can derive the estimated initial 6DOF camera pose $\pi^{\prime}$.
With the assistance of methods such as \cite{zhou2021nerd, li2022symmnerf}, we can further refine our 6DOF camera pose estimation.



\subsubsection{Camera Pose and BA-NeRF}
It is important to note that assuming object symmetry introduces a significant constraint; if the object is not symmetrical, the flipped images and their inferred camera poses can destabilize the NeRF model. To mitigate this, we perform BA-based NeRF \cite{lin2021barf} with estimated initial camera poses for refining flipped camera poses and learning 3D space simultaneously.

When training the model with the flipped images' camera poses initialized randomly, the learning process tends to proceed in an incorrect direction. However, by using our 6DOF pose estimation, it becomes suitable for BA-based NeRF, which requires moderately estimated camera poses. Considering our confidence in the initial input pose, the training strategy we adopted ensures that only the flipped pose undergoes bundle adjustment, while the original remains unaffected. We can formulate homogeneous coordinates of pixel coordinates as $\overline{\mathbf{p}} = [\mathbf{p}; 1]^{T}$. Then, we can represent a 3D point $\mathbf{x}$ on the viewing ray at depth $z$ by using the equation $\mathbf{x} = z \overline{\mathbf{p}}$. Given a camera pose $\pi^{\prime}$, 3D point $\mathbf{x}$, within the camera's view space undergoes a transformation to align with the world coordinates using a rigid 3D transformation $\mathcal{W}$. Consequently, the RGB value rendered at a particular pixel, intrinsically dependent on the camera pose, is illustrated as the following:
\begin{align} 
    \nonumber
    \hat{I^{\prime}}(\mathbf{p} ; \pi^{\prime})=\mathcal{R}\left(F_{\Theta}\left(\mathcal{W}\left(z_1 \overline{\mathbf{p}} ; \pi^{\prime}\right) \right), \ldots, F_{\Theta}\left(\mathcal{W}\left(z_N \overline{\mathbf{p}} ; \pi^{\prime}\right) \right)\right) \text{.}
\end{align}
Here, our goal is to optimize flipped camera poses $\Pi^{\prime}=\{\pi_i^{\prime}, ... , \pi_n^{\prime}\}$ and network parameters $\Theta$. Consequently, using Eqn.~\eqref{eqn:redefined_nerf_loss} we can have the optimal solution as:
\begin{align} 
    \nonumber
    {\Pi^{\prime}}^{opt}, {\Theta}^{opt} =
    \argmin_{\Pi^{\prime}, \Theta}
    \sum_{i=1}^N \sum_{\mathbf{p}}\left\|I_i^{\prime}(\mathbf{p})-\hat{I^{\prime}}\left(\mathbf{p} ; \pi_i^{\prime}, \Theta\right)\right\|_2^2 \text{.}
\end{align}
Given the inherent non-linearity of the optimization problem, we employed a gradient-based optimization approach. We implemented the Lucas-Kanade algorithm used in Optical Flow and leveraging the Jacobian, conducted backpropagation. 

\subsection{NeRF with uncertainty}



We started our approach based on the assumption of object symmetry, attempting to predict unobserved views by flipping images. However, not all objects exhibit symmetry, and depending on the viewing angle of an object, a simple flip can result in performance decrease. We handle this issue in the aspect of camera pose estimation, but similar challenge occurs in images as well.
Since various objects are not perfectly symmetric, in Eqn.~\eqref{eqn:rendering function} some of rays $\mathbf{r}$ gathered from the flipped images $I^\prime$ gives us significant errors while training NeRF.
To ensure robust predictions of unobserved views by ignoring wrong pixels and exploiting correct pixels from $I^\prime$, we propose a training method that takes into account the uncertainty associated with flipped images. Following similar approach to the bundle adjustment on the flipped pose, we consider uncertainty values solely for the flipped images.

To address the uncertainty of the predicted color value, we follow Bayesian framework \cite{kendall2017uncertainties}.
To compute the loss for observed images $I$ in Eqn.~\eqref{eqn:redefined_nerf_loss}, 
\begin{align}
\label{c-c}
    L(\mathbf{r})={\left\|\mathbf{C}_i(\mathbf{r})-\hat{\mathbf{C}}(\mathbf{r})\right\|_2^2}
\end{align}
In Eqn.~\eqref{c-c} all rays $\mathbf{r}$ in $I$ are fully exploited for training NeRF, which matters in our case, since 3D objects are not perfectly symmetric thereby some of the rays $\mathbf{r}^\prime$ from $I^\prime$ are not matched to the 3D shape.
Instead of representing the ray (or color value) from any given point in the scene as a singular value, we propose modeling it using a Gaussian distribution for the observation uncertainty. This predicted variance can be interpreted as an indicator of the data's uncertainty for a specific location. Consequently, the mapping function can be represented as $F_\Theta : (\mathbf{x},\mathbf{d}) \rightarrow (\beta^{2}, \mathbf{c}, \sigma)$, 
where $\beta^{2}$ is the variance of color (uncertainty).
To validate the variance value, the Softplus function $\bar{\beta^{2}} = \beta_{min}^{2} + \log(1 + exp(\beta^{2}(\mathbf{r}(t))))$ was adopted. Contrary to the flipped images, the original input image set the value of $\bar{\beta^{2}}$ to $\beta_{min}^{2}$. This approach compels the model to exhibit heightened uncertainty in unobserved regions, thereby facilitating effective predictions of unobserved views. 
Based on Eqn.~\eqref{eqn:rendering function}, the rendering function $\mathcal{R}$, being a linear combination of sampled points, leads the rendered color value $\hat{C}(\mathbf{r})$ to follow a Gaussian distribution. Same way, we can derive the variance of rendered color $\hat{\beta}^2(\mathbf{r})$. Building on this, the loss for the ray $\mathbf{r}^\prime$ in flipped image $I^\prime$, taking into account uncertainty, is defined as follows:
\begin{align} 
    \nonumber    L(\mathbf{r}^\prime)=\frac{\left\|\mathbf{C}_i(\mathbf{r}^\prime)-\hat{\mathbf{C}}(\mathbf{r}^\prime)\right\|_2^2}{2 \hat{\beta}^2(\mathbf{r})}+\frac{\log \hat{\beta}^2(\mathbf{r}^\prime)}{2}+\frac{w}{N_s} \sum_{j=1}^{N_s} \sigma \left(t_j\right) \text{.}
\end{align}
The above cost function aims to minimize the negative log-likelihood, with the final term serving as a regularization term to prevent blurring effects.

We initialized the flipped camera pose using a geometric approach. Given the unreliability of the flipped information, we refined the camera pose, taking into account the uncertainty associated with the image pixels. Our method does not require an additional network, making it both efficient and rapid, and particularly apt for robotics applications. In the subsequent section, we compare our proposed method with the baseline method. \\

\section{Experiments}

\subsection{Implementation and Setup}

\begin{figure}[hbt!]
    \centering
    \includegraphics[width=6.0cm, height=4.7cm]{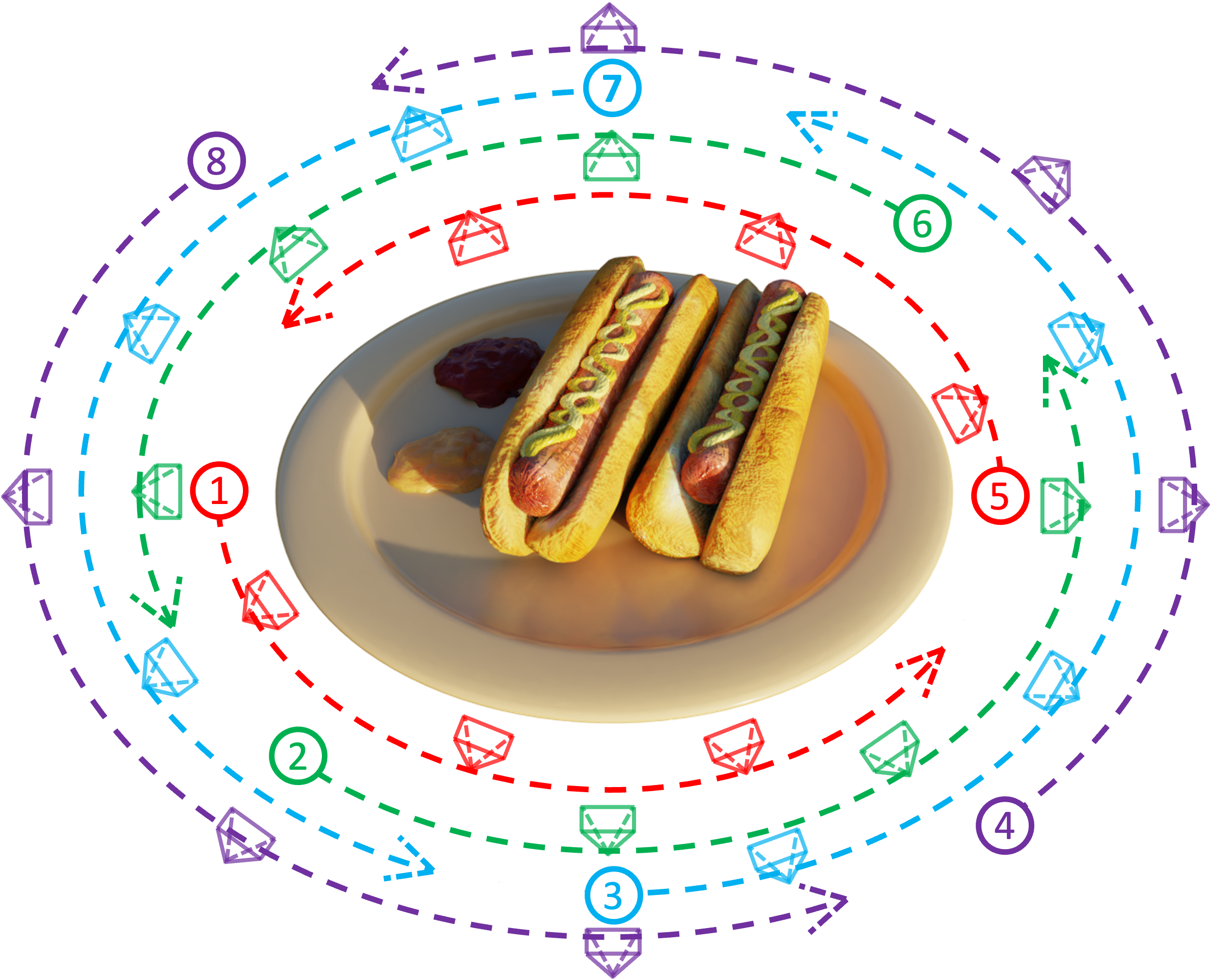}
    \caption{Dataset configurations. To generalize performance across different views, we generate a total of eight dataset configurations for a single data scene. For each configuration, we selected 16 evenly spaced images by rotating around an object. The $n^{th}$ dataset configuration encompasses images from the $(2n - 1)^{th}$ to the $(2n + 6)^{th}$. Through this setup, we demonstrate our method can effectively predict unobserved views, not just in specific views but in more general ones.}
    \label{8_dataset}
    \vspace{-10pt}
\end{figure}


Given that we possess the ground-truth for unobserved views, we used the NeRF synthetic dataset \cite{mildenhall2021nerf} for our experimentation. This dataset comprises objects that exhibit a certain degree of structural symmetry (Chairs, Lego, Materials, Ship) and those without such symmetrical properties (Drums, Ficus, Hotdog, Mic). Neither of these cases has the symmetry of lighting and color.
Considering the input data scenario, as shown in Fig.~\ref{overview_image} we assume that the robot
is limited to observing only partial side of the object. 
In alignment with real-world robot applications, we trained with a limited image set, i.e., in our case eight RGB images. Since flipping an image can lead to varying performance depending on the observation view, we deemed it essential to ensure a comprehensive evaluation of general performance. Therefore, we constructed a total of eight different input dataset configurations, as illustrated in Fig.~\ref{8_dataset}. We conducted comparative experiments on five distinct methods (baseline, baseline + uncertainty, flip, flip + uncertainty, upper bound), which will be discussed in detail in the analysis. For simplicity in all our experiments, we employed a single network that samples 128 points. Considering that uncertainty during training can lead to instability, thus we perform conventional NeRF training up to 10k epochs and apply Bayesian approach up to 50k epochs subsequently. The remaining configurations were kept consistent with the vanilla NeRF setup, all experiments were conducted on NVIDIA RTX 4090 GPU. As our approach strives to predict the view that cannot be observed, we utilize the unobserved view from the opposite side as the validation and ground-truth image. For quantitative comparison, we used three representative metrics: Peak Signal-to-Noise Ratio (PSNR) \cite{korhonen2012peak}, Structural Similarity Index Measure (SSIM) \cite{yue2005similarity}, and Learned Perceptual Image Patch Similarity (LPIPS) \cite{zhang2018unreasonable}.

\subsection{Analysis of the experiment results}



First, we provide a description of the five methods in detail. Our dataset consists of four sets: the input image set obtained from the robot exploration, the flipped image set, the test image set, and the upper image set for the unexplored scene. Each set contains eight images, and we assume that the corresponding camera poses are already known except for the camera pose of the flipped image. The input image set consists of data that the robot has acquired while navigating, whereas the flipped image set is created by flipping the input images.
\begin{table}[t]
    \caption{Summary of the five experimental methods.}
    \label{summarize 5 method}
    \begin{tabular}{l|l}
    \hline
        (a) \textit{(B/L)}                     & input image\\ \hline
        (b) \textit{(B/L) + U}                 & input image + consider ucert \\ \hline
        (c) \textit{Flip}                     & input image + flipped and refine pose \\ \hline
        (d) \textit{Flip + U}                 & input image + flipped and refine pose + consider ucert \\ \hline
        (e) \textit{Upper}                  & input image + ground-truth image   \\ \hline
    \end{tabular}
\end{table}
\begin{table}[t]
    \centering
    \caption{ablation study. We report PSNR, SSIM, and LPIPS of the full model(the last row) and three configurations by removing the application of bundle adjustment, uncertainty on the flip image and both, respectively.}
    \label{ablation}
    \begin{tabular}{ccccc}
    \toprule
    bundle adjustment & uncertainty & PSNR$\uparrow$ & SSIM$\uparrow$ & LPIPS$\downarrow$ \\ \midrule
          -            &      -       & 16.636   & 0.787  & 0.232   \\
    \checkmark        &        -     & 17.205   & 0.789  & 0.230   \\
          -            & \checkmark  & 16.694   & 0.790  & 0.240   \\
    \checkmark        & \checkmark  & \textbf{17.620}   & \textbf{0.794}  & \textbf{0.229}   \\ \bottomrule
    \vspace{-10pt}
    \end{tabular}
\end{table}
The test set comprises unobserved views that the robot cannot explore. We construct the upper data by selecting views that are closest to those in the test data. We summarize the five methods in Table.~\ref{summarize 5 method}. (a) \textit{baseline (B/L)} is trained using the image set obtained while the robot is navigating. (b) \textit{B/L + uncertainty (U)} is applied uncertainty to all images used in the baseline. (c) \textit{Flip} leverages both the input image and its flipped version, and utilizes the least squares method to estimate the flipped camera pose for training. Moreover, it refines the pose of the flipped image through bundle adjustment. Our proposed method, (d) \textit{Flip + U}, extends (c) \textit{Flip} by additionally considering the uncertainty of the flipped image. Meanwhile, (e) \textit{Upper} method trains the model using both the input image, upper image, and corresponding upper camera pose. This approach is an upper bound since it directly uses the image of the unobserved view in training. To examine each element of our proposed method, we conducted an ablation study and show the results in Table.~\ref{ablation}. The results indicated that applying both bundle adjustment and uncertainty to the flipped image yielded the best performance.
%
%
 \begin{table*}[t]
    \centering
    \caption{Quantitative comparisons of our flip methods with baseline method and upper bound method on the NeRF synthetic dataset \cite{mildenhall2021nerf}. The upper image is a case where an image observed from a viewpoint that the robot has not explored is added to the image used at the baseline. The experiments, as depicted in \ref{8_dataset}, were conducted using a total of 8 input dataset configurations for each scene, and the metric value averaged out. Our method demonstrates performance improvement in predicting unobserved views, even without utilizing areas the robot has not explored for training. (* B/L mean baseline and U mean uncertainty)}
    \label{experiment_table}
    \begin{tabular}{c|ccccc|ccccc|ccccc}
    \hline
    \multirow{2}{*}{Scene} & \multicolumn{5}{c|}{PSNR $\uparrow$} & \multicolumn{5}{c|}{SSIM $\uparrow$} & \multicolumn{5}{c}{LPIPS $\downarrow$}                           \\ \cline{2-16} 
                     & \textit{B/L} & \textit{B/L+U} & \textit{Flip} & \textit{Flip+U} & \textit{Upper}          & \textit{B/L} & \textit{B/L+U} & \textit{Flip}    & \textit{Flip+U} & \textit{Upper}              & \textit{B/L} & \textit{B/L+U} & \textit{Flip} & \textit{Flip+U} & \textit{Upper}    \\ \hline
    Chairs          & 14.07  & 16.46 & 19.69 & \textbf{19.86} & 26.96         & 0.75 & 0.78 & 0.87 & \textbf{0.88} & 0.92          & 0.48 & 0.28 & 0.13 & \textbf{0.12} & 0.05 \\
    Lego            & 11.57  & 14.85 & 15.24 & \textbf{17.79} & 23.91         & 0.65 & 0.67 & 0.74 & \textbf{0.78} & 0.87          & 0.50 & 0.34 & 0.23 & \textbf{0.20} & 0.11 \\
    Materials       &  9.66  & 16.74 & 17.29 & \textbf{19.14} & 25.60         & 0.66 & 0.75 & 0.81 & \textbf{0.83} & 0.90          & 0.50 & 0.31 & 0.18 & \textbf{0.14} & 0.09 \\
    Ship            &  6.73  & 16.97 & 18.29 & \textbf{18.70} & 24.77         & 0.55 & 0.65 & 0.71 & \textbf{0.72} & 0.80          & 0.61 & 0.40 & 0.27 & \textbf{0.26} & 0.18 \\
    Drum            & 10.24  & 11.91 & 11.90 & \textbf{13.77} & 20.19         & 0.60 & 0.59 & 0.63 & \textbf{0.68} & 0.83          & 0.55 & 0.43 & 0.47 & \textbf{0.43} & 0.16 \\
    Ficus           & 11.20  & 15.59 & 17.15 & \textbf{18.02} & 22.59         & 0.72 & 0.78 & 0.81 & \textbf{0.82} & 0.88          & 0.33 & 0.22 & 0.15 & \textbf{0.13} & 0.10 \\
    Hotdog          & 13.67  & \textbf{19.07} & 18.91 & 18.92 & 28.30         & 0.70 & 0.83 & 0.84 & \textbf{0.84} & 0.93          & 0.44 & 0.22 & \textbf{0.19} & 0.21 & 0.07 \\
    Mic             & 12.89  & 14.16 & 13.78 & \textbf{14.30} & 24.93         & 0.82 & 0.78 & \textbf{0.82} & 0.80 & 0.93          & 0.46 & 0.30 & 0.38 & \textbf{0.30} & 0.07 \\
    \hline
    Mean            & 11.25  & 15.71 & 16.53 & \textbf{17.56} & 24.65         & 0.68 & 0.72 & 0.77 & \textbf{0.79} & 0.88          & 0.48 & 0.30 & 0.25 & \textbf{0.22} & 0.10\\ \hline

    \end{tabular}
\end{table*}
\begin{figure*}[h]
    \centering 
    \includegraphics[width=17.4cm, height=10.0cm]{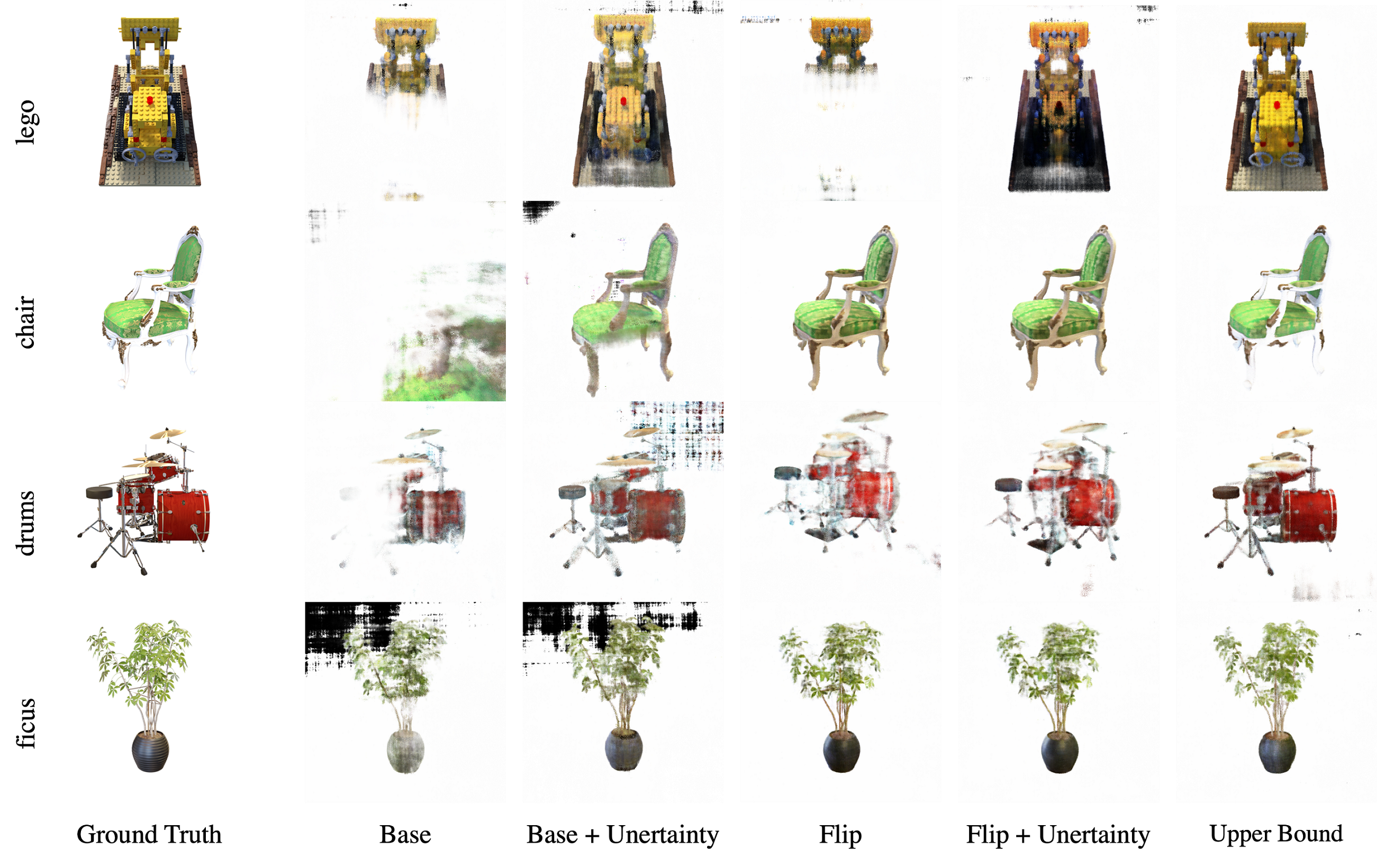} 
    \caption{Qualitative comparisons on the NeRF synthetic dataset \cite{mildenhall2021nerf}. Our method captures the object structure of unobserved views effectively, offering a more accurate image than other methods while maintaining a noise-free output.}
    \label{Qualitative comparisons}
    \vspace{-10pt}
\end{figure*}

In TABLE \ref{experiment_table}, we display the metric results of the five methods. Naturally, for objects with inherent structural symmetry, flipping the image from certain views can effectively predict unobserved views. To validate performance not just in specific scenarios but also in more general settings, we conducted experiments across eight dataset configurations as depicted in \ref{8_dataset}, presenting the results as an average. The primary objective of our experiment is to predict unobserved views. Therefore, (a) \textit{B/L}, which relies solely on one side of an object, exhibits significantly low performance. As shown in Fig. \ref{Qualitative comparisons}, certain portions disappear, and in specific views, the resulting image is entirely blank. (b) \textit{B/L + U}, which considers the uncertainty of the input image, demonstrates performance improvements even without flipping the image. However, as evidenced in the qualitative results, (b) \textit{B/L + U} has an issue with black noise appearing in unobserved views. (c) \textit{Flip}, on the other hand, does show an enhancement in performance compared to (c) \textit{B/L} and effectively addresses the black noise issue. Yet, for non-symmetric objects, its performance lags behind that of (b) \textit{B/L + U}. Furthermore, as demonstrated in the lego scene of \ref{Qualitative comparisons}, configurations that observe only the front face of an object do not always benefit from the flip method. Contrasting this with the chair scene, where the side of the object is observed, there are instances where the unobserved views aren't accurately predicted despite flipping. By integrating the advantages of both (b) \textit{B/L + U} and (c) \textit{Flip}, our proposed approach (d) \textit{Flip + U} demonstrates significant performance improvements, regardless of the object's symmetry. \\

\section{Conclusion}


In this paper, we introduced a flipped observation generation method for NeRF to predict unobserved views. Most of the robot exploration is performed in real-time, thus it is challenging to observe all the possible viewpoints. To predict unobserved views, we propose a method that exploits the observed views by image flipping. Given that flipped images do not have camera poses, we also proposed a method for estimating the 6DOF pose of the flipped images.
Furthermore, we applied bundle adjustment to optimize these estimated poses in conjunction with the model's parameters.
Since only the observed parts should be exploited in order to have robust unobserved view prediction, for the flipped images, we incorporated Bayesian approach by considering the uncertainty estimation. From experimental results we show that our approach markedly improves performance in predicting scenes from unobserved viewpoints and thus shows competitive results compared to the traditional approaches. Our flipped observation method is suitable for robotic applications such as robot exploration and navigation where real-time is important since our approach is simple but robust.


\bibliographystyle{IEEEtran}
\bibliography{bibmy}

\end{document}